\documentclass[a4paper]{cas-dc}
\pdfoutput=1
\usepackage[numbers]{natbib}
\usepackage{tablefootnote}
\usepackage{amsmath,amssymb,amsfonts,bm}
\usepackage[T1]{fontenc}
\usepackage{booktabs}
\usepackage{url}

\usepackage{graphicx}
\usepackage{caption}
\usepackage{microtype}
\usepackage{bm}
\usepackage{enumitem}
\usepackage{chngcntr}
\counterwithin{figure}{section}

\newcommand{\forceindent}{\leavevmode{\parindent=2em\indent}}
\begin{document}

\shorttitle{LightX3ECG}
\shortauthors{Khiem et al.}

\title[mode = title]{LightX3ECG: A Lightweight and eXplainable Deep Learning System for 3-lead Electrocardiogram Classification}

\author[1,2]{\color{black} Khiem H. Le } \cormark[1]
\author[1,2]{\color{black} Hieu H. Pham }
\author[1,2]{\color{black} Thao BT. Nguyen }
\author[1,2]{\color{black} Tu A. Nguyen }
\author[3]{\\\color{black} Tien N. Thanh }
\author[1,2]{\color{black} Cuong D. Do }
\begin{NoHyper}
\cortext[cor1]{Corresponding author: \url{khiem.lh@vinuni.edu.vn}}
\end{NoHyper}
\address[a]{College of Engineering and Computer Science, VinUniversity, Ha Noi, Viet Nam}
\address[b]{VinUni-Illinois Smart Health Center, VinUniversity, Ha Noi, Viet Nam}
\address[c]{College of Health Sciences, VinUniversity, Ha Noi, Viet Nam}



\begin{abstract}
Cardiovascular diseases (CVDs) are a group of heart and blood vessel disorders that is one of the most serious dangers to human health, and the number of such patients is still growing. Early and accurate detection plays a key role in successful treatment and intervention. Electrocardiogram (ECG) is the gold standard for identifying a variety of cardiovascular abnormalities. In clinical practices and most of the current research, standard 12-lead ECG is mainly used. However, using a lower number of leads can make ECG more prevalent as it can be conveniently recorded by portable or wearable devices. In this research, we develop a novel deep learning system to accurately identify multiple cardiovascular abnormalities by using only three ECG leads. Specifically, we use three separate One-dimensional Convolutional Neural Networks (1D-CNNs) as backbones to extract features from three input ECG leads separately. The architecture of 1D-CNNs is redesigned for high performance and low computational cost. A novel Lead-wise Attention module is then introduced to aggregate outputs from these three backbones, resulting in a more robust representation which is then passed through a Fully-Connected (FC) layer to perform classification. Moreover, to make the system’s prediction clinically explainable, the Grad-CAM technique is modified to produce a high meaningful lead-wise explanation. Finally, we employ a pruning technique to reduce system size, forcing it suitable for deployment on hardware-constrained platforms. The proposed lightweight, explainable system is named LightX3ECG. We got classification performance in terms of F1 scores of 0.9718 and 0.8004 on two large-scale ECG datasets, i.e., Chapman and CPSC-2018, respectively, which surpassed current state-of-the-art methods while achieving higher computational and storage efficiency. Visual examinations and a sanity check were also performed to demonstrate the strength of our system’s interpretability. To encourage further development, our source code is publicly available at \url{https://github.com/lhkhiem28/LightX3ECG}. 
\end{abstract}

\maketitle

\section{Introduction}
\begin{figure}[ht]
    \centering
    \includegraphics[keepaspectratio, height=225.7pt]{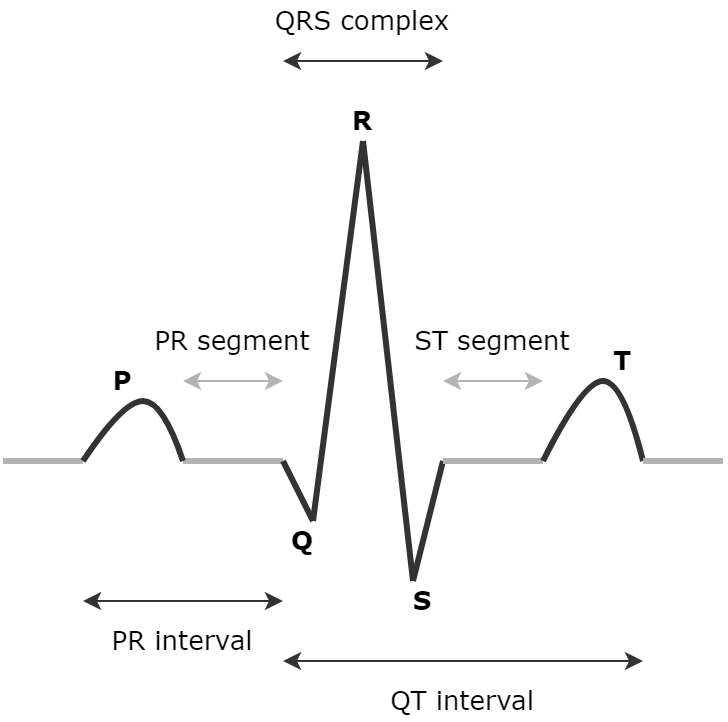}
    \caption{The usual structure of an ECG signal.}
    \label{fig:ECG}
\end{figure}

\forceindent
Cardiovascular diseases (CVDs) are one of the primary sources of death globally, accounting for 17.9 million deaths in 2019, representing 32\% of all deaths worldwide. Also, three-quarters of these deaths take place in low- and middle-income countries, according to World Health Organization \footnote{\url{https://www.who.int/health-topics/cardiovascular-diseases}}. Therefore, it's critical to detect these heart problems as soon as possible so that treatment may begin with counseling and medications. Electrocardiogram (ECG) is a waveform representation of the electrical activity of the heart obtained by placing electrodes on the body surface. The usual structure of an ECG signal \cite{Electrocardiography}, as illustrated in Figure \ref{fig:ECG}, consists of three main components: P wave, which represents depolarization of atria; QRS complex, which represents depolarization of ventricles; and T wave, which represents repolarization of ventricles. Other parts of the signal include PR, QT intervals, or PR, ST segments. This electrical signal is a widely used, non-invasive tool for identifying cardiovascular abnormalities in patients. However, ECG analysis is a professional and time-consuming task, it requires cardiologists with a high degree of training to carefully examine and recognize pathological patterns in ECG recordings. This challenge, coupled with the rapid increase in ECG data, makes computer-aided, automatic ECG analysis more and more essential, especially in low- and middle-income countries, where high-quality and experienced cardiologists are extremely scarce. 
\\\\
The 12-lead ECG, which is standard for hospital and clinic usage, is typically recorded from electrodes placed on the patient's limbs and on the surface of the chest. Thus, twelve ECG leads can be broken down into two main types: six limb leads (I, II, III, aVR, aVL, aVF) and six chest leads (V1, V2, V3, V4, V5, V6). Conventional 12-lead ECG has been also demonstrated to be effective for a variety of ECG analysis tasks by many previous efforts \cite{Cardiology_Challenge_2020, Study_12-lead_ECG_data, Deep_Learning_ECG_Review}. Acquiring 12-lead ECG, on the other hand, is redundant and heavily relied on clinical equipment with limited accessibility. Recently, breakthroughs in ECG technologies have led to the development of smaller, lower-cost, and easier-to-use ECG-enabled devices \cite{Zio_Patch, Digital_Stethoscope, Apple_Watch}. These advancements have paved the way for point-of-care screening and continuous monitoring using signals recorded by these devices \cite{Point-of-care_screening, remote_ECG_monitoring}. However, these types of devices only produce a subset of standard twelve leads, sometimes even just one lead. This raises the need for building ECG analysis methods that only rely on this subset of leads rather than the entire set. For that reason, in this study, we use a combination of only three ECG leads (I, II, and V1) as input for our system to strike a balance between high classification performance and the ease of signal acquisition. Leads I and II are used because they are easy to acquire and favored by cardiologists for quick review, as well as represent relatively enough information for six limb leads, according to some laws and equations \cite{Handbook}. Lead V1 is used to incorporate information about chest leads into the input. 
\\
Existing approaches for automatic ECG analysis can be divided into two categories: traditional methods and deep learning-based methods. In traditional methods, which are also known as two-stage methods, human experts hand-craft meaningful features from raw ECG signals such as statistical features (e.g., mean, standard deviation, variance, and percentile) or time- and frequency-domain features, referred to as expert features \cite{Feature_Extraction_ECG_Survey, Feature_Extraction_ECG}. Then, these features are concatenated and fed into some kinds of machine learning algorithms. The performance of these methods is significantly limited by the quality of expert features and the capability of machine learning algorithms applied. The second approach is to use end-to-end deep learning models that offer a high model capability without the need for domain knowledge and an explicit feature extraction stage \cite{current_state-of-the-art}. These types of models have gained significant improvements compared to the former approach \cite{Deep_Learning_ECG_Review}. Deep learning models have dramatically improved the state-of-the-art in speech recognition, visual object recognition, object detection, and many other areas such as drug discovery and genomics \cite{Deep_Learning}. Despite their superior performance, deep learning models are plagued by two well-known drawbacks: their black-box nature and increasingly large model size which limit their applicability in real-world scenarios. In this study, we aim to design an accurate ECG classification system that also overcomes these two issues. 
\\\\
In almost of previous works on deep learning-based 12-lead ECG classification, all twelve leads are standardized to the same length, then vertically stacked together to form a unified input and fed into a followed deep learning model \cite{Deep_Learning_ECG_Review, Study_12-lead_ECG_data}. This strategy works well when dealing with 12-lead ECG. However, when dealing with a smaller number of leads, such as three, we propose to use three distinct models as separate backbones to handle three input ECG leads separately, which will be demonstrated in this study to give us better performance. This multi-input strategy is reasonable since these kinds of signals usually require separate treatment. In more detail, we employ three distinct redesigned One-dimensional Squeeze-and-Excitation Residual Networks (1D-SEResNets) \cite{Squeeze-and-Excitation}, which are highly effective for dealing with ECG data, to extract features from three input signals. Then, inspired by the attention mechanism \cite{Attention, Attention_Survey}, we design a novel Lead-wise Attention module as our aggregation technique to explore the most essential input lead and merge outputs of these backbones, resulting in a more robust representation that is then sent through an FC layer to perform classification. 
\\\\
Although deep learning models can achieve state-of-the-art performance in a range of predictive tasks, they are often viewed as black boxes. Due to high complexity, predictions made by these models are not traceable by humans. In many applications, especially in the medical domain, understanding the model's behavior is as important as the accuracy of its predictions since it is difficult for cardiologists or pathologists to accept unexplainable decisions \cite{AI_medicine_explainable}. This makes Explainable AI (XAI) become a highly active research topic in the past few years \cite{Explainable_AI_Survey}. In this study, we also construct an XAI framework for our 3-lead ECG classification task using class activation maps. Our XAI technique called Lead-wise Grad-CAM provides three different class activation maps for three input ECG leads, giving more clinical interpretability to our system. Another disadvantage of deep learning models, as previously discussed, is the expansion in model size. The majority of existing ECG classification research are primarily concerned with enhancing classification performance while paying little attention to model size, leading to memory-intensive models that are impractical for hardware-constrained platforms deployment \cite{mobile_deep_learning_perspective}. To improve the proposed system’s suitability for point-of-care screening and remote monitoring deployment on these platforms, we apply a pruning technique to make the system lightweight and easy to distribute while just slightly sacrificing its performance. 
\\\\
To summarize, our main contributions are as follows: 
\begin{figure*}
    \centering
    \includegraphics[keepaspectratio, height=273.7pt]{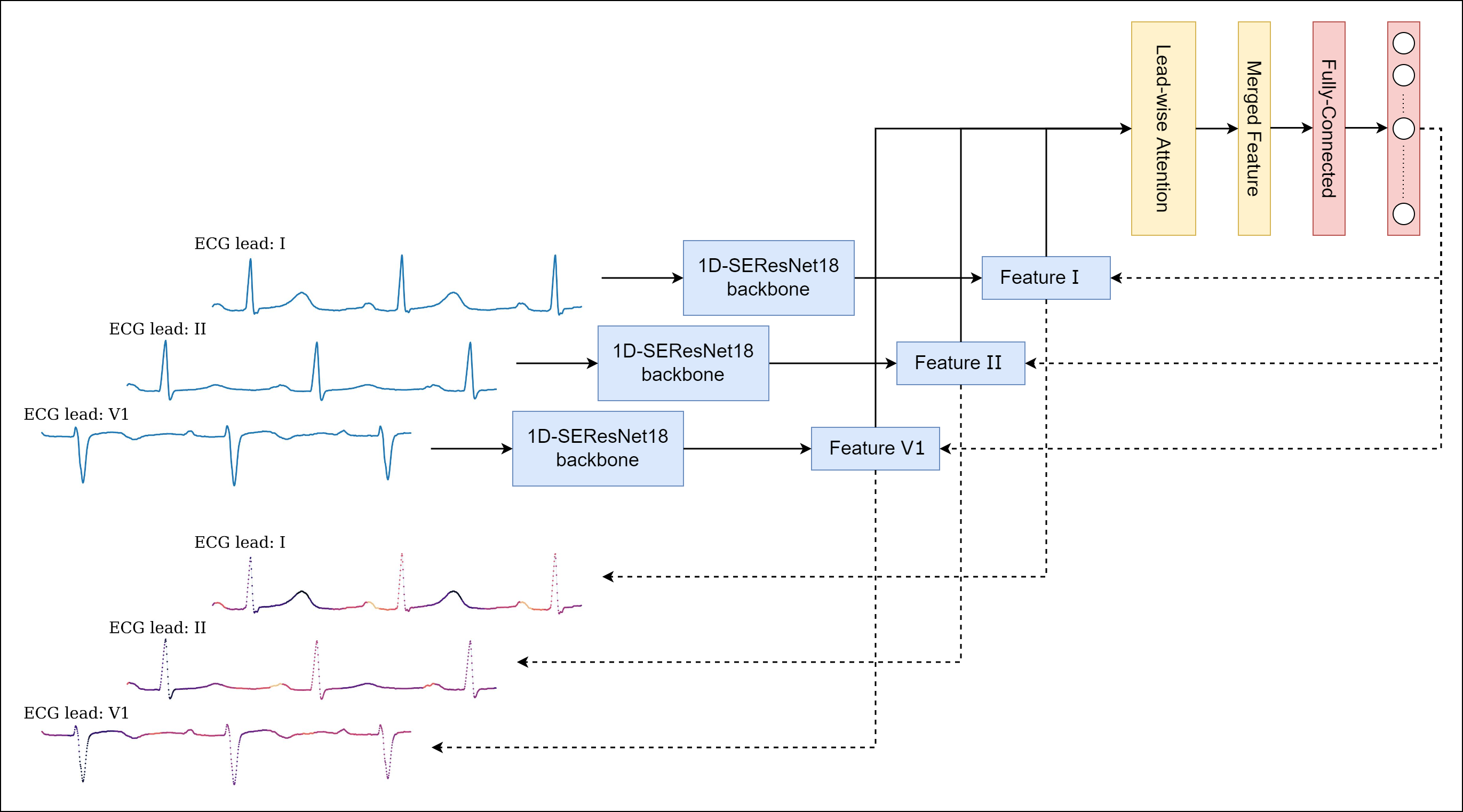}
    \caption{An overview of the proposed system. Dashed arrows indicate the interpreting stage.}
    \label{fig:System}
\end{figure*}
\begin{itemize}[leftmargin=*]
\item We propose an accurate deep learning system for 3-lead ECG classification which consists of three redesigned 1D-SEResNet backbones followed by a novel Lead-wise Attention module and an FC layer, as shown in Figure \ref{fig:System}. 
\item A novel XAI technique named Lead-wise Grad-CAM is introduced, which is adapted from the common Grad-CAM technique on the system’s architecture, giving a better explanation for the made prediction. 
\item We further employ a pruning technique to reduce the system’s space on memory while mostly preserving its classification performance. 
\item Extensive experiments were conducted on two large-scale multi-lead ECG datasets, i.e., Chapman and CPSC-2018 where our system showed superior performance in both multi-class and multi-label classification manners while enhancing compactness and clinical interpretability. 
\end{itemize}
The rest of this article is organized as follows. Section 2 provides a survey of literature related to our work. In Section 3, we present components of the proposed system in detail. Section 4 describes the experimental setup and our results. An ablation study is performed in Section 5. Finally, we give further discussions and conclude this work in Section 6. 

\section{Related Work}
\forceindent
In this section, we discuss some research directions and existing works that are highly related to our work, including deep learning-based ECG analysis, reduced-lead ECG classification, and explainable AI for ECG classification. 
\\\\
\textbf{Deep learning-based ECG Analysis}. Deep learning-based methods have been the preferred approach for ECG analysis over the last few years \cite{current_state-of-the-art}. Specifically, 1D-CNNs have become popular when dealing with ECG data because of their one-dimension structure. Acharya et al. \cite{9_layer_CNN_5_classes} early developed a 9-layer 1D-CNN to identify 5 different types of cardiovascular abnormalities. Recently, researchers have begun to use more sophisticated 1D-CNN architectures, particularly ones whose 2D version achieves high image classification accuracy. Zhang et al. \cite{GradientSHAP} proposed using 1D-ResNet34, Zhu et al. \cite{34_layer_CNN_27_classes} ensembled two 1D-SEResNet34s and one set of expert rules to identify more types of abnormalities. Furthermore, Yao et al. \cite{TI-CNN} constructed Time-Incremental ResNet18 (TI-ResNet18), a combination of a 1D-ResNet18 and an LSTM network in order to capture both spatial and temporal patterns in ECG signals. Other than CVD detection, deep learning models have been employed on ECG data for other variety of tasks. Li et al. \cite{Sleep_Apnea} combined a sparse Autoencoder and Hidden Markov Model for diagnosing obstructive sleep apnea. Moreover, Santamaria-Granados et al. \cite{Emotion_Detection} focused on emotion, classifying the affective state of a person. Attia et al. \cite{Drug_Assessment} performed a proof of concept study on non-invasive drug assessment based on ECG signals. Rahman et al. \cite{COV-ECGNET, hexaxial_feature_mapping} tried to early diagnose COVID-19 using ECG trace images. 
\\\\
\textbf{Reduced-lead ECG Classification}. In recent years, some small, low-cost, and easy-to-use ECG-enable devices have been introduced in the market \cite{Zio_Patch, Digital_Stethoscope, Apple_Watch}. These devices are different from clinical equipment in that they only provide a subset of standard twelve ECG leads, sometimes just one. Thus, in most cases, newer methods are being developed to do ECG classification based on single- or reduced-lead data rather than standard 12-lead data. While single-lead ECG is currently limiting in performance, early studies have suggested that reduced-lead ECG may hold potential. Drew et al. \cite{myocardial_ischemia} demonstrated that interpolated 12-lead ECG, which is derived from a reduced-lead set (limb leads plus V1 and V5), is comparable to standard 12-lead ECG for diagnosing wide-QRS-complex tachycardias and acute myocardial ischemia. Green et al. \cite{coronary_syndrome} also found that the leads III, aVL, and V2 together yielded a similar performance as the full 12-lead ECG for diagnosing acute coronary syndrome. Cho et al. \cite{myocardial_infarction} claimed that myocardial infarction could be detected not only with a conventional 12-lead ECG but also with a limb 6-lead ECG. Our work provides further support to demonstrate the ability of reduced-lead ECG for identifying a range of cardiovascular abnormalities, not just a few. 
\\\\
\textbf{Explainable AI for ECG Classification}. While the black-box nature of deep learning models may be ignorable in many contexts, it leads to a lack of responsibility and trust in decisions made in sensitive areas like medicine and healthcare. Hence, researchers have started to bring popular XAI techniques applied to image data into ECG data. Hughes et al. \cite{LIME} proposed to use of Linear Interpretable Model-Agnostic Explanations (LIME). Zhang et al. \cite{GradientSHAP}, Anand et al. \cite{XAI_cardiac_disorders} applied SHapley Additive exPlanations (SHAP) analysis to test the interpretability of an ECG classification model. LIME and SHAP are both perturbation-based techniques that provide explanations based on the variation of output after applying perturbations to input. Some disadvantages of these techniques are combinatorial complexity explosion and producing explanations by very concrete class activation maps \cite{Perturbation-based}. Due to inherent smoothing in provided explanations, some XAI techniques such as Grad-CAM and its variants are recently more preferred. Vijayarangan et al. \cite{Single-Lead_ECG_Grad-CAM}, Raza et al. \cite {federated_transfer_learning} employed Grad-CAM on 1D-CNN for single-lead ECG classification. Ganeshkumar et al. \cite{Multi-Lead_ECG_Grad-CAM} further applied Grad-CAM on a multi-lead circumstance but generated the same class activation map for multiple input signals. In this work, we leverage the system’s architecture with the multi-input strategy and our Lead-wise Attention module to subtlely adapt Grad-CAM and provide one different informative class activation map for each of three input ECG leads. 
\section{Proposed System}
\begin{figure*}
    \centering
    \includegraphics[keepaspectratio, height=273.7pt]{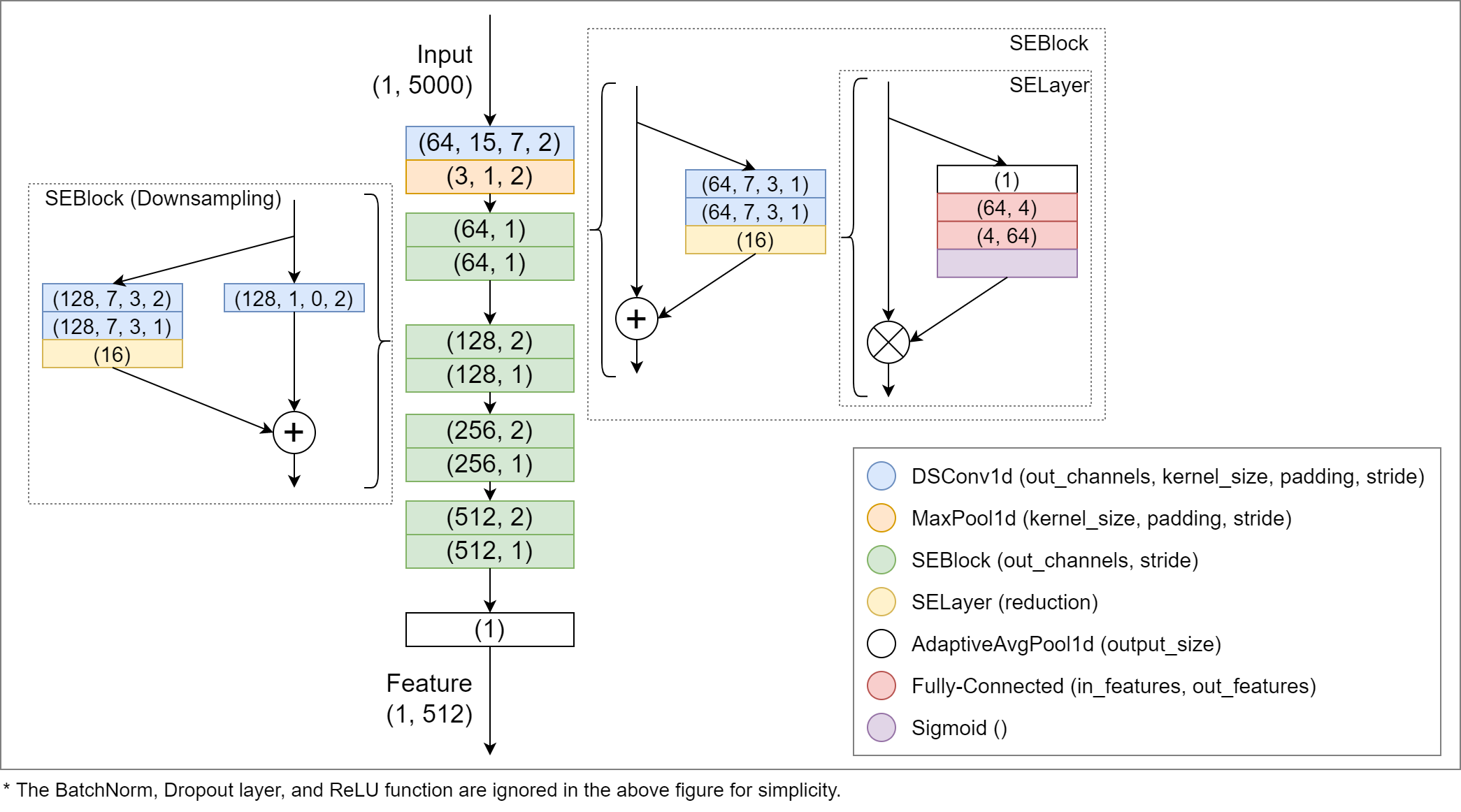}
    \caption{The architecture of 1D-SEResNet backbones.}
    \label{fig:Backbones}
\end{figure*}
\forceindent
In this section, we present the whole proposed system in detail. Firstly, the architecture of 1D-SEResNet backbones is described. Next, we sequentially introduce our novel Lead-wise Attention module and XAI technique, Lead-wise Grad-CAM. The pruning technique, which is used to establish LightX3ECG, is briefly discussed last. An overview of our LightX3ECG is shown in Figure \ref{fig:System}. 

\subsection{1D-SEResNet Backbones}
To achieve high performance and low computational cost backbones, we redesign 1D-SEResNet18 \cite{Squeeze-and-Excitation}, which consists of 18 main layers, in two steps as follows. 
\\\\
First, Convolution (Conv) layers are modified with a much larger kernel size to expand those receptive fields in order to capture longer patterns in ECG signals. This strategy has been suggested more effective for ECG data in specific \cite{ECG-signal-deep_CNN-and-BiLSTM}, and time-series data in general \cite{InceptionTime}. Second, we replace all of the standard Conv layers with Depth-wise Separable Conv (DSConv) layers for reducing the number of parameters of the model. Introduced in MobileNets \cite{MobileNetV2, MobileNetV3}, DSConv splits computation of standard Conv into two parts. The first part is depth-wise, in which each filter only convolutes each input channel. Another part is point-wise, using a 1x1 filter to combine multi-channel outputs of depth-wise layers. This design reduces the total number of parameters of our system by 80\%. This architecture is used for all three backbones and is illustrated in Figure \ref{fig:Backbones}. 

\subsection{Lead-wise Attention}
\begin{figure}
    \centering
    \includegraphics[keepaspectratio, width=238pt]{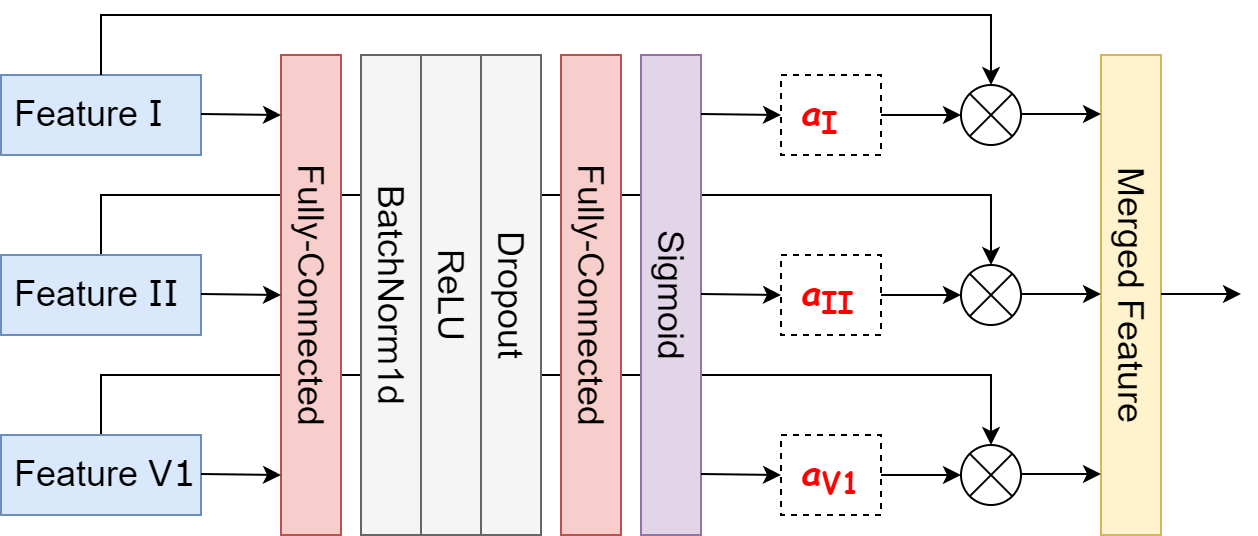}
    \caption{The proposed Lead-wise Attention module.}
    \label{fig:Lead-wise Attention}
\end{figure}

To achieve an end-to-end classification system, the outputs, also known as features or embeddings, extracted from backbones, must be combined. Typically, one can combine these features by simply applying a summation or concatenation operation to them, but this is usually ineffective due to their simplicity. Inspired by the success of the attention mechanism in many areas \cite{Attention_Survey}, we propose a Lead-wise Attention module to more effectively ensemble these features together and acquire a final robust feature which is then routed to the last FC layer, the classifier, to perform classification. Our Lead-wise Attention module is described in Figure \ref{fig:Lead-wise Attention}. 
\\\\
Firstly, features from backbones are concatenated and sent through a sequential list of layers including an FC, a BatchNorm, a Dropout, followed by another FC layer and a Sigmoid function to determine the attention score, or importance score for each feature. Subsequently, the final feature is obtained by taking a weighted sum over these features by corresponding generated scores. This module can be formulated: 
\begin{gather*}
\text{f}_{\text{merged}}=\sum_{i=1}^{3}\boldsymbol{\alpha}_{i}\text{f}_{i}, \\\boldsymbol{\alpha}=\texttt{Sigmoid}(\texttt{FC}(\texttt{FC}(\texttt{Concat}[\text{f}_{i}|i=\overline{1,3}]))).
\end{gather*}

\subsection{Lead-wise Grad-CAM}
In the domain of imaging, Grad-CAM \cite{Grad-CAM} is one of the most famous techniques to provide interpretability to 2D-CNNs which uses values of gradients flowing into the final Conv layer to produce a class activation map (CAM) \cite{chest_X-ray, blood_smear_images}. CAM is a heatmap that highlights class-specific regions of an image which the model looked at to classify that image. In this work, we adapt Grad-CAM to our system for the same aim, which we refer to as Lead-wise Grad-CAM, in the following steps. 
\\\\
First, similar to standard Grad-CAM, we employ values of gradients flowing into the final Conv layers of three backbones to gather three distinct CAMs $C_{i,i=\overline{1,3}}$ corresponding to three input ECG leads. In addition to CAMs provided by Grad-CAM, the proposed system has an additional source of interpretability, the importance scores $\boldsymbol{\alpha}_{i,i=\overline{1,3}}$ gathered from the Lead-wise Attention module that show the contribution of each backbone’s feature to the prediction of the system, therefore, show the contribution of each input signal. To take advantage of this insight from our Lead-wise Attention module, we multiply three CAMs by corresponding importance scores to get more informative heatmaps. Finally, for visualization, these heatmaps are normalized and overlaid on corresponding input ECG lead: 
\begin{equation*}
M_{i}=\texttt{normalize}(\boldsymbol{\alpha}_{i}C_{i})
\end{equation*}

\subsection{Pruning}
A deep learning-based method often involves a large model and massive computation. Hence, when operating the proposed system on portable or wearable devices, issues such as insufficient memory or computational resources are noticeable. As a direct solution, we apply the weights pruning \cite{Learning-Compression, Weights_and_Connections} technique to compress the system and make it can be executed completely on these devices. 
\\\\
Weights pruning is a post-training model compression technique to make a trained model more sparse. This is accomplished by increasing the number of zero-valued elements present in the model's weights. In this work, we prune 80\% weights of the system with the lowest $L_{1}$-norm in order to reduce the system’s space on memory 3 times while mostly maintaining its classification performance, and finally establishing LightX3ECG as a result. The idea is that weights with small $L_{1}$-norm, or absolute value, contribute little to the prediction of the system, so they are less important and can be zeroed out. 

\section{Experiments and Results}
\forceindent
In this section, we comprehensively describe our study design and all experimental results. Two datasets and implementation details are introduced first, then we report the performance of LightX3ECG and its interpretability. 

\subsection{Datasets}
To benchmark the performance of the proposed system, we conducted experiments on two of the largest public real-world datasets for ECG classification, i.e., Chapman and CPSC-2018. Diagnosis class frequency and patient characteristics of these two datasets are shown in Table \ref{tab:datasets}. 
\\\\
\textit{Chapman} \cite{Chapman}. Chapman University and Shaoxing People's Hospital collaborated to establish this large-scale \textit{multi-class} ECG dataset which consisted of 10.646 12-lead ECG recordings collected from 10.646 patients. Each recording was taken over 10 seconds with a sampling rate of 500 Hz and labeled with 11 common diagnostic classes. The amplitude unit was microvolt. These 11 classes were grouped into 4 categories including AFIB, GSVT, SB, and SR. AFIB consists of atrial fibrillation and atrial flutter, GSVT contains supraventricular tachycardia, atrial tachycardia, atrioventricular node reentrant tachycardia, atrioventricular reentrant tachycardia, and sinus atrium to atrial wandering rhythm, SB only includes sinus bradycardia, and SR includes sinus rhythm and sinus irregularity. 
\\
\textit{CPSC-2018} \cite{CPSC-2018}. In 2018, the first China Physiological Signal Challenge organized during the 7th International Conference on Biomedical Engineering and Biotechnology released a publicly available large-scale \textit{multi-label} ECG dataset. This dataset contained 6.877 12-lead ECG recordings with a sampling rate of 500 Hz and durations ranging from 6 to 60 seconds. Millivolt was the amplitude unit. These ECG recordings were labeled with 9 diagnostic classes including SNR (normal sinus rhythm), AF (atrial fibrillation), IAVB (first-degree atrioventricular block), LBBB (left bundle branch block), RBBB (right bundle branch block), PAC (premature atrial contraction), PVC (premature ventricular contraction), STD (ST-segment depression), STE (ST-segment elevation). 

\begin{table}[ht]
\setlength{\tabcolsep}{6.60pt}
\caption{Description of two datasets.\\Mean and standard deviation are reported for age.}
\begin{tabular}{lrrc}
\midrule
\multicolumn{4}{l}{Chapman} \\
\midrule
Class & Frequency (\%) & Male (\%) & Age \\
\midrule
AFIB & 2225 (20.90) & 1298 (58.34) & 72.90 $\pm$ 11.68 \\
GSVT & 2307 (21.67) & 1152 (49.93) & 55.44 $\pm$ 20.49 \\
SB   & 3889 (36.53) & 2481 (63.80) & 58.34 $\pm$ 13.95 \\
SR   & 2225 (20.90) & 1025 (46.07) & 50.84 $\pm$ 19.25 \\
\midrule
\\
\midrule
\multicolumn{4}{l}{CPSC-2018} \\
\midrule
Class & Frequency (\%) & Male (\%) & Age \\
\midrule
SRN  & 918  (13.35) & 363  (39.54) & 41.56 $\pm$ 18.45 \\
AF   & 1221 (17.75) & 692  (56.67) & 71.47 $\pm$ 12.53 \\
IAVB & 722  (10.50) & 490  (67.87) & 66.97 $\pm$ 15.67 \\
LBBB & 236  (03.43) & 117  (49.58) & 70.48 $\pm$ 12.55 \\
RBBB & 1857 (27.00) & 1203 (64.78) & 62.84 $\pm$ 17.07 \\
PAC  & 616  (08.96) & 328  (53.25) & 66.56 $\pm$ 17.71 \\
PVC  & 700  (10.18) & 357  (51.00) & 58.37 $\pm$ 17.90 \\
STD  & 869  (12.64) & 252  (29.00) & 54.61 $\pm$ 17.49 \\
STE  & 220  (03.20) & 180  (81.82) & 52.32 $\pm$ 19.77 \\
\midrule
\end{tabular}
\label{tab:datasets}
\end{table}
\subsection{Implementation Details}
To ensure the reproducibility of our results, the experimental setup including the data preprocessing step, training setting, and evaluation strategy, are all described. 
\\\\
\textit{Data Preprocessing}: As a deep learning system requires inputs to be of the same length, all ECG recordings were fixed at 10 seconds in length in both datasets. This was done by truncating the part exceeding the first 10 seconds for longer recordings and padding shorter ones with zero. We took leads I, II, and V1 from each ECG recording to construct the input with the shape of 3x5000 and fed it into our system. 
\\
\textit{Data Augmentation}: To reach a better generalization, we additionally propose the DropLead augmentation technique which randomly dropped one of three input signals with a probability of 50\% during training. This was accomplished by masking the selected signal with all of zero. DropLead is not applied during the inference stage. 
\\
\textit{Training and Evaluation}: For evaluation, we applied a 10-fold cross-validation strategy following some previous works \cite{Chapman, GradientSHAP}. We stratify divided each of the two datasets into 10 folds and performed 10 rounds of training and evaluation. At each round, 8 folds; 1 fold; and 1 remaining fold were used as training, validation, and test set, respectively. In the multi-label classification manner, the optimal threshold of each class was searched in a range (0.05, 0.95) with a step of 0.05 to achieve the best F1 score on the validation set. We report the average performance of 10 rounds on the test set in terms of precision, recall, F1 score, and accuracy. For training, the proposed system was optimized by Adam optimizer \cite{Adam} with an initial learning rate of 1e-3 and a weight decay of 5e-5 for 70 epochs. We used Cosine Annealing scheduler \cite{SGDR} in the first 40 epochs to reschedule the learning rate to 1e-4 and then kept it constant in the last 30 epochs. Cross-entropy and binary cross-entropy were utilized as loss functions in multi-class and multi-label manners, respectively. Finally, after weights pruning was applied, our system was fine-tuned for 5 epochs with the same setting except the learning rate was held constant at 1e-4. All experiments were run on a machine with an NVIDIA GeForce RTX 3090 TURBO 24G. 

\begin{table}[ht]
\setlength{\tabcolsep}{8.05pt}
\caption{Performance detail of LightX3ECG on two datasets.}
\begin{tabular}{lrrrr}
\midrule
\multicolumn{5}{l}{Chapman} \\
\midrule
Class & Precision & Recall & F1 score & Accuracy \\
\midrule
AFIB & 0.9750 & 0.9662 & 0.9706 & 0.9878 \\
GSVT & 0.9510 & 0.9612 & 0.9561 & 0.9807 \\
SB   & 0.9823 & 0.9987 & 0.9904 & 0.9930 \\
SR   & 0.9860 & 0.9550 & 0.9703 & 0.9878 \\
\midrule
Average & 0.9736 & 0.9703 & 0.9718 & 0.9873 \\
\midrule
\\
\midrule
\multicolumn{5}{l}{CPSC-2018} \\
\midrule
Class & Precision & Recall & F1 score & Accuracy \\
\midrule
SRN  & 0.6903 & 0.8342 & 0.7554 & 0.9266 \\
AF   & 0.9344 & 0.9461 & 0.9402 & 0.9789 \\
IAVB & 0.9014 & 0.8828 & 0.8920 & 0.9775 \\
LBBB & 0.9038 & 0.8704 & 0.8868 & 0.9913 \\
RBBB & 0.9454 & 0.9428 & 0.9441 & 0.9702 \\
PAC  & 0.6972 & 0.5758 & 0.6307 & 0.9353 \\
PVC  & 0.8796 & 0.7197 & 0.7917 & 0.9637 \\
STD  & 0.7870 & 0.7824 & 0.7847 & 0.9469 \\
STE  & 0.6486 & 0.5217 & 0.5783 & 0.9746 \\
\midrule
Average & 0.8209 & 0.7862 & 0.8004 & 0.9628 \\
\midrule
\end{tabular}
\label{tab:performance}
\end{table}

\subsection{System Performance}

\begin{table*}[!ht]
\setlength{\tabcolsep}{7.76pt}
\caption{Comparison of the proposed system to other methods.}
\label{tab:comparison}
\begin{tabular}{l|rrrrr}
\midrule
Method & F1 on Chapman & F1 on CPSC-2018 & No. Params (M) & No. FLOPs (B) & Size (MB) \\
\midrule
1D-ResNet34 \cite{GradientSHAP}   & 0.9624 & 0.7684 & 16.61 & 5.91 & 58.18 \\
1D-SEResNet34 \cite{34_layer_CNN_27_classes} & 0.9659 & 0.7845 & 16.76 & 5.91 & 58.75 \\
TI-ResNet18 \cite{TI-CNN}  & 0.9647 & 0.7872 & 11.39 & 1.42 & 40.51 \\
InceptionTime \cite{InceptionTime} & 0.9417 & 0.7352 & \textbf{0.45} & 2.29 & \textbf{1.63} \\
LightX3ECG (Ours)   & \textbf{0.9718} & \textbf{0.8004} & 5.31 & \textbf{1.34} & 6.52 \\
LightX3ECG (w/o pruning)   & 0.9722 & 0.8010 & 5.31 & 1.34 & 19.28 \\
\midrule
\end{tabular}
\end{table*}

We got F1 scores of 0.9718 and 0.8004 on two datasets, i.e., Chapman and CPSC-2018, respectively. Overall, accuracy for each class exceeded 0.92 and the average exceeded 0.96 in both. However, we also observed that F1 scores of PAC and STE classes were limited, which could be due to the insufficiency of these diagnosis classes in the CPSC-2018 dataset. Detailed performance is presented in Table \ref{tab:performance}. 
\\
For precisely benchmarking, we compared LightX3ECG with some popular ECG classification methods, which can be considered state-of-the-art including 1D-ResNet34 \cite{GradientSHAP}, 1D-SEResNet34 \cite{34_layer_CNN_27_classes}, InceptionTime \cite{InceptionTime}, and TI-ResNet18 \cite{TI-CNN}. For fair comparisons, all of these methods were implemented and trained using 3-lead ECG as input and setting similar to our system. Comparisons of F1 scores, complexity, and compactness are shown in Table \ref{tab:comparison}. LightX3ECG outperformed other methods in both datasets while achieving the lowest computational cost with FLOPs at 1.34B. In terms of storage efficiency, our system only took up 6.52MB on disk, which was much less than the other three methods. Additionally, the performance of the system without applying weights pruning showed that effectively using this technique helps reduce the system’s space significantly with a negligible side-effect. 

\subsection{System Interpretability} \label{4.4}
A comprehensive validation was conducted to demonstrate LightX3ECG’s interpretability, including a visual check and a methodical check. 

\subsubsection{Visual examinations}
\begin{figure}[ht]
    \centering
    \includegraphics[keepaspectratio, width=226.77pt]{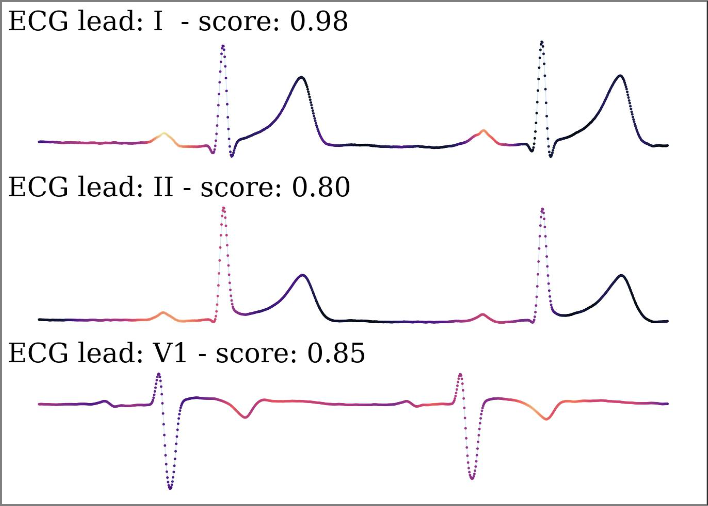}
    \caption{The explanation for a sample SNR ECG recording.}
    \label{fig:SNR}
\end{figure}

\begin{figure}[ht]
    \centering
    \includegraphics[keepaspectratio, width=226.77pt]{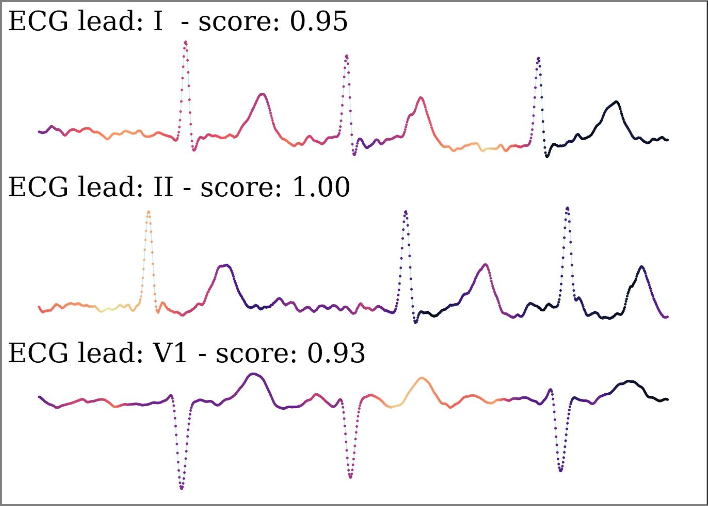}
    \caption{The explanation for a sample AF ECG recording.}
    \label{fig:AF}
\end{figure}

\begin{figure}[ht]
    \centering
    \includegraphics[keepaspectratio, width=226.77pt]{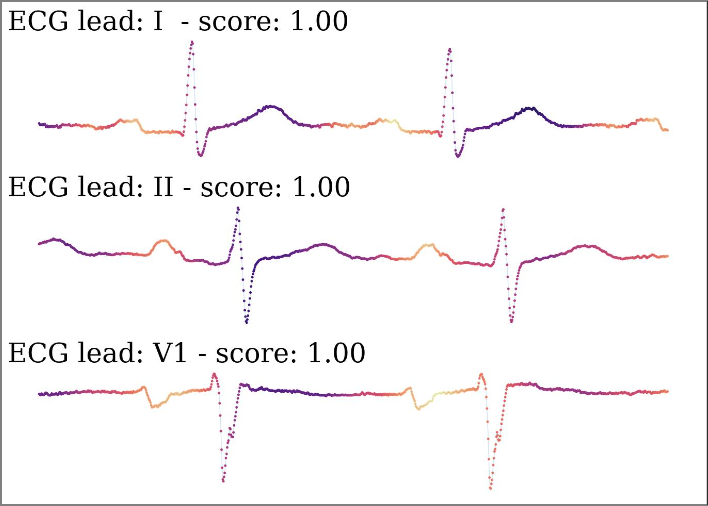}
    \caption{The explanation for a sample IAVB ECG recording.}
    \label{fig:IAVB}
\end{figure}

\begin{figure}[ht]
    \centering
    \includegraphics[keepaspectratio, width=226.77pt]{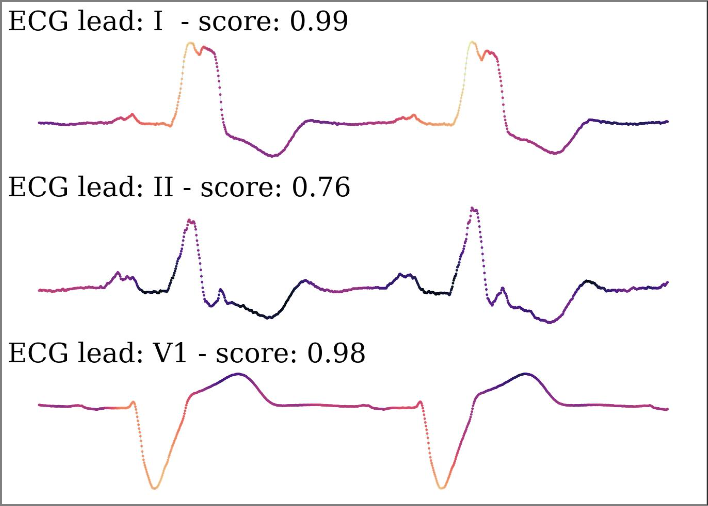}
    \caption{The explanation for a sample LBBB ECG recording.}
    \label{fig:LBBB}
\end{figure}

\begin{figure}[ht]
    \centering
    \includegraphics[keepaspectratio, width=226.77pt]{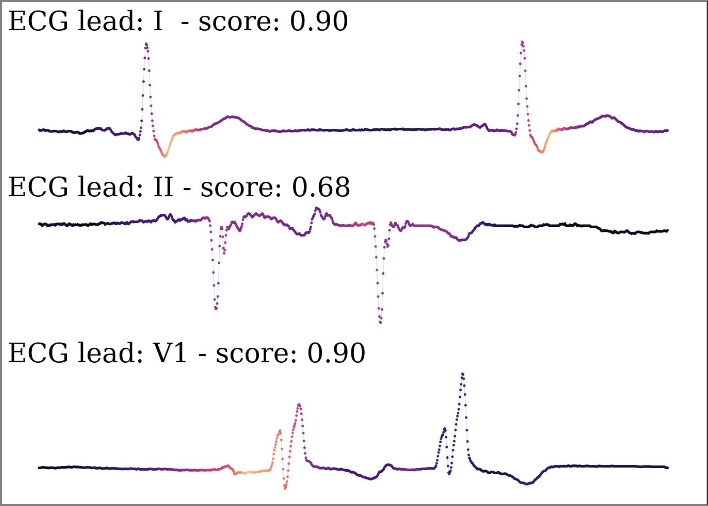}
    \caption{The explanation for a sample RBBB ECG recording.}
    \label{fig:RBBB}
\end{figure}

\begin{figure}[ht]
    \centering
    \includegraphics[keepaspectratio, width=226.77pt]{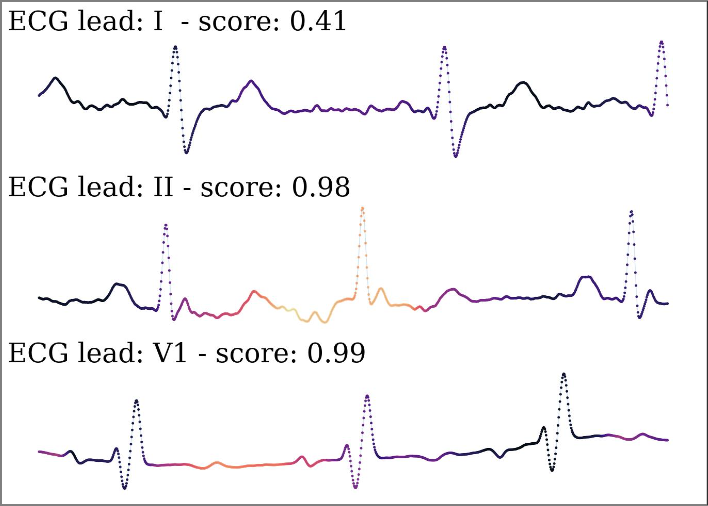}
    \caption{The explanation for a sample PAC ECG recording.}
    \label{fig:PAC}
\end{figure}

\begin{figure}[ht]
    \centering
    \includegraphics[keepaspectratio, width=226.77pt]{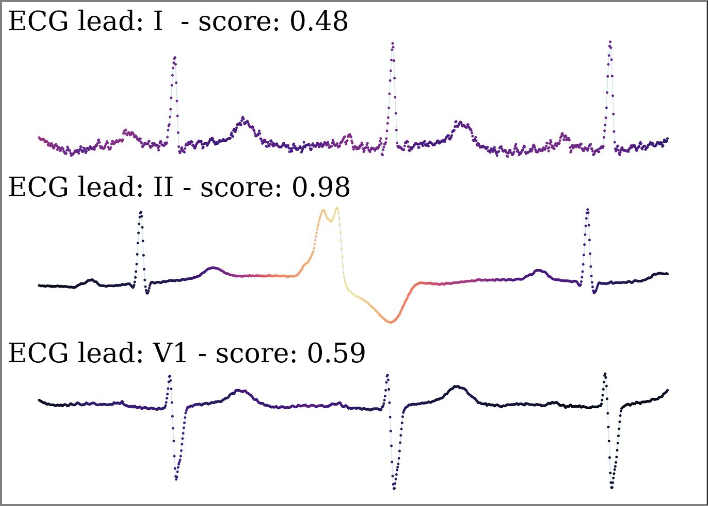}
    \caption{The explanation for a sample PVC ECG recording.}
    \label{fig:PVC}
\end{figure}

\begin{figure}[ht]
    \centering
    \includegraphics[keepaspectratio, width=226.77pt]{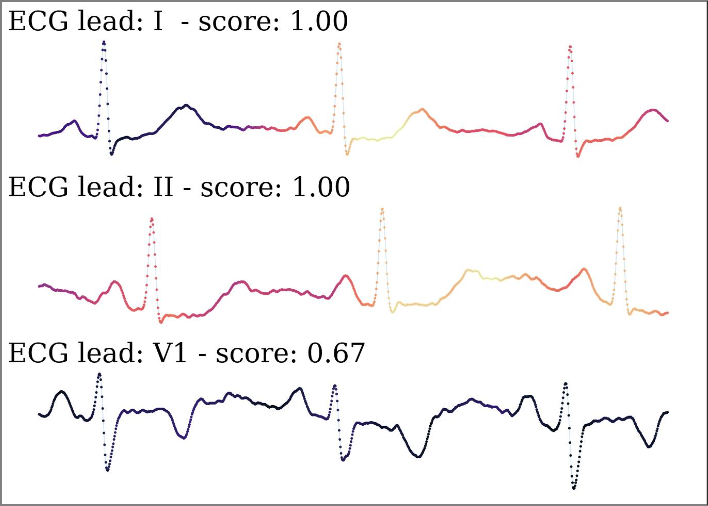}
    \caption{The explanation for a sample STD ECG recording.}
    \label{fig:STD}
\end{figure}

\begin{figure}[ht]
    \centering
    \includegraphics[keepaspectratio, width=226.77pt]{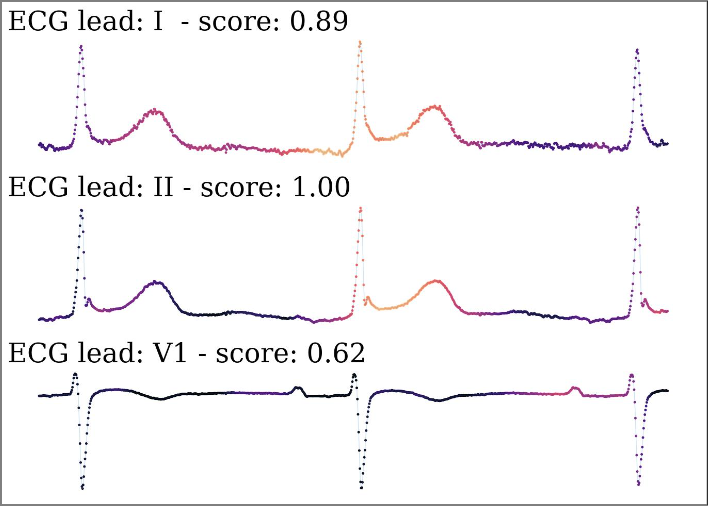}
    \caption{The explanation for a sample STE ECG recording.}
    \label{fig:STE}
\end{figure}

For visual check, we carefully reviewed the explanation from the system for a sample ECG recording belonging to each of the diagnosis classes in the CPSC-2018 dataset and compared it with some cardiological evidence collected from a variety of sources \cite{Gutierrez, Chou, IAVB, Goldberger, Alventosa, Surawicz}, and LITFL ECG Library \footnote{\url{https://litfl.com/category/ecg-library}}. 
\\\\
\textit{1) SNR (normal sinus rhythm)}. An SNR ECG recording has a normal P wave preceding each QRS complex, which is also standard, as seen in Figure \ref{fig:ECG}. Also, P waves upright in leads I and II. From activation maps in Figure \ref{fig:SNR}, we can see that system strongly focused on regions of P waves in leads I and II. Thus, the explanation is consistent with the diagnostic criteria of SNR. The importance scores indicated that lead I contributed more to the system's prediction than others. 
\\
\textit{2) AF (atrial fibrillation)}. An AF ECG recording has irregular QRS complexes with the lack of P waves. Also, fibrillatory waves are usually visible in lead V1. From activation maps in Figure \ref{fig:AF}, we can see that system recognized the lack of P waves in leads I and II, and fibrillatory waves in lead V1. Thus, the explanation is consistent with the diagnostic criteria of AF. The importance scores indicated that three leads contributed roughly equally to the system's prediction. 
\\
\textit{3) IAVB (first-degree atrioventricular block)}. An IAVB ECG recording has prolonged PR intervals. Also, P waves are buried in the preceding T wave. From activation maps in Figure \ref{fig:IAVB}, we can see that system recognized the prolonged PR intervals in leads I and II. Thus, the explanation is consistent with the diagnostic criteria of IAVB. The importance scores indicated that three leads contributed roughly equally to the system's prediction. 
\\
\textit{4) LBBB (left bundle branch block)}. An LBBB ECG recording has broad QRS complexes. Also, S waves are fairly deep in lead V1. From activation maps in Figure \ref{fig:LBBB}, we can see that system recognized broad QRS complexes in lead I, and deep S waves in lead V1. Thus, the explanation is consistent with the diagnostic criteria of LBBB. The importance scores indicated that leads I and V1 mostly contributed to the system's prediction. 
\\
\textit{5) RBBB (right bundle branch block)}. An RBBB ECG recording has wide slur S waves in lead I. Also, “M-shaped” QRS complexes are visible in lead V1. From activation maps in Figure \ref{fig:RBBB}, we can see that system recognized wide slur S waves in lead I, and “M-shaped” QRS complexes in lead V1. Thus, the explanation is consistent with the diagnostic criteria of RBBB. The importance scores indicated that leads I and V1 mostly contributed to the system's prediction. 
\\
\textit{6) PAC (premature atrial contraction)}. A PAC ECG recording has abnormal (non-sinus) P waves followed by a normal QRS complex. Also, P waves are usually negative in lead II. From activation maps in Figure \ref{fig:PAC}, we can see that system recognized non-sinus P waves in leads II and V1, specifically negative P waves in lead II. Thus, the explanation is consistent with the diagnostic criteria of PAC. The importance scores indicated that leads II and V1 mostly contributed to the system's prediction. 
\\
\textit{7) PVC (premature ventricular contraction)}. A PVC ECG recording has some sporadic periods that are abnormal compared to surrounding periods. Also, QRS complexes in these periods are irregular too. From activation maps in Figure \ref{fig:PVC}, we can see that system recognized abnormal periods compared to surrounding periods in lead II and irregular QRS complexes in these periods. Thus, the explanation is consistent with the diagnostic criteria of PVC. The importance scores indicated that lead II contributed more to the system's prediction than others. 
\\
\textit{8) STD (ST-segment depression)}. As its name, an STD ECG recording has depressed ST segments. From activation maps in Figure \ref{fig:STD}, we can see that system recognized depressed ST segments in leads I and II. Thus, the explanation is consistent with the diagnostic criteria of STD. The importance scores indicated that leads I and II mostly contributed to the system's prediction. 
\\
\textit{9) STE (ST-segment elevation)}. As its name, an STE ECG recording has elevated ST segments. From activation maps in Figure \ref{fig:STE}, we can see that system recognized elevated ST segments in leads I and II. Thus, the explanation is consistent with the diagnostic criteria of STE. The importance scores indicated that leads I and II mostly contributed to the system's prediction. 

\subsubsection{Sanity check}
Recent works in the literature on XAI research have strongly emphasized the importance of implementing sanity checks \cite{Sanity_Checks} in order to assess the quality of XAI techniques methodically \cite{roadmap_XAI, Sensitivity}. These types of checks verify whether or not the provided explanation is related to the model's parameters or the data used for training. 

\begin{figure}[t]
    \centering
    \includegraphics[keepaspectratio, width=226.77pt]{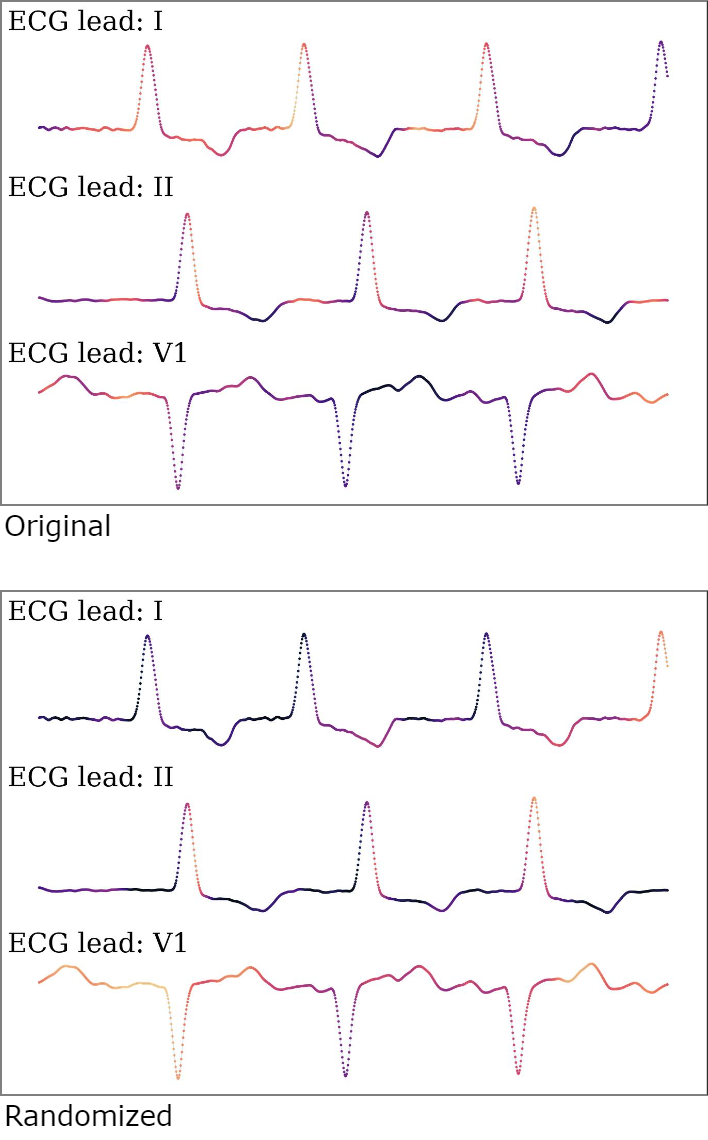}
    \caption{An example of a comparison between explanations from the original system and the randomized system.}
    \label{fig:Sanity}
\end{figure}

For this purpose, we performed a simple parameter randomization test, which is one of two forms of sanity checks, to assess our Lead-wise GradCAM technique. In particular, by using Lead-wise GradCAM, we compared explanations for a hundred ECG recordings from our trained system (original system) with those from the randomized system. the randomized system was accomplished by perturbing the final FC layer, classifier, of the original system. Figure \ref{fig:Sanity} shows an example of this comparison, as we expect, explanations differ. We also report the average Spearman’s rank correlation of these explanations in Table \ref{tab:correlation}. Lead-wise GradCAM and SHAP analysis \cite{GradientSHAP} both passed this sanity check, but our technique gave a lower correlation score.

\begin{table}[ht]
\setlength{\tabcolsep}{7.7pt}
\caption{Spearman's rank correlation of explanations\\between the original system and randomized system.}
\begin{tabular}{lrr}
\midrule
\multicolumn{1}{l|}{Method} & Chapman & CPSC-2018 \\
\midrule
\multicolumn{1}{l|}{SHAP \cite{GradientSHAP}}   & 0.16 & 0.18 \\
\multicolumn{1}{l|}{Lead-wise Grad-CAM (Ours)} & \textbf{0.10} & \textbf{0.11} \\
\midrule
\end{tabular}
\label{tab:correlation}
\end{table}

\section{Ablation Study}
\forceindent
In this section, we conducted an ablation study to validate the effect of three selected input leads on the performance of LightX3ECG. In particular, we fixed leads I and II as our limb leads while substituting lead V1 with another chest lead to create a new combination of three leads that we used as input for the system. Table \ref{tab:ablation} shows that the performance was fairly consistent among combinations and the combination of leads (I, II, and V1) produced the best performance in both datasets by a slight margin. 

\begin{table*}
\setlength{\tabcolsep}{8.84pt}
\caption{Comparison with different chest leads on the performance of LightX3ECG.}
\begin{tabular}{l|cccccc}
\midrule
\multicolumn{1}{l|}{Dataset} & (I, II, and V1) & (I, II, and V2) & (I, II, and V3) & (I, II, and V4) & (I, II, and V5) & (I, II, and V6) \\
\midrule
Chapman   & \textbf{0.9718} & 0.9702  & 0.9705  & 0.9702  & 0.9714  & 0.9711 \\
CPSC-2018 & \textbf{0.8004} & 0.8002  & 0.7997  & 0.7959  & 0.8001  & 0.7992 \\
\midrule
\end{tabular}
\label{tab:ablation}
\end{table*}

\section{Discussions \& Conclusion}
\forceindent
In this article, we introduced an efficient and accurate deep learning system that uses a combination of three 10-second ECG leads (I, II, and V1) to identify cardiovascular abnormalities. We posed a new state-of-the-art for the 3-lead ECG classification task, where the proposed system outperformed most of the existing methods available for ECG classification in terms of F1 scores, complexity, and compactness. Additionally, we focused heavily on the XAI framework, which can give a more meaningful and clinical explanation for the system's prediction, making it more valuable in medical contexts. Our system was also compressed to be ready for the production stage. However, the multi-input architecture of LightX3ECG is not suitable for small datasets and leads to difficulty in training on large-scale datasets, as well as a high storage cost which needs a practical technique like weights pruning to overcome. In the future, LightX3ECG will be improved to identify wider varieties of cardiovascular abnormalities, as well as be generalized on different sources of data. Demographic data such as age and gender will be incorporated to boost current performance. And a novel XAI framework for this multi-modal input will be also developed. 

\bibliography{refs}
\bibliographystyle{unsrt}

\appendix
\section{Appendix} \label{appendix}
\forceindent
In this section, we provide supplementary figures for Subsection \ref{4.4}, including more examples for visual examinations, related to Figures 4. 1-9, and more examples for the sanity check, related to Figure 4.10. 

\subsection{Visual examinations}
\newpage
\textit{1) SNR (normal sinus rhythm)}
\begin{figure}[ht]
    \centering
    \includegraphics[keepaspectratio, width=225.53pt]{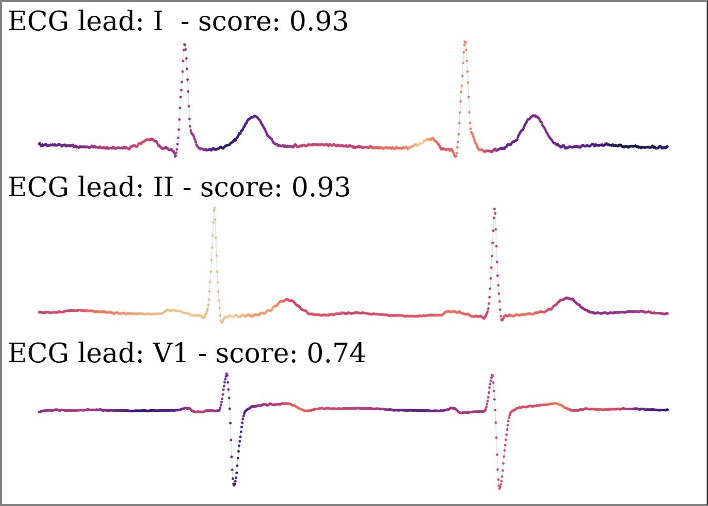}
    \caption{The explanation for another SNR ECG recording.}
    \label{fig:SNR_2}
\end{figure}
\\
\textit{2) AF (atrial fibrillation)}
\begin{figure}[ht]
    \centering
    \includegraphics[keepaspectratio, width=225.53pt]{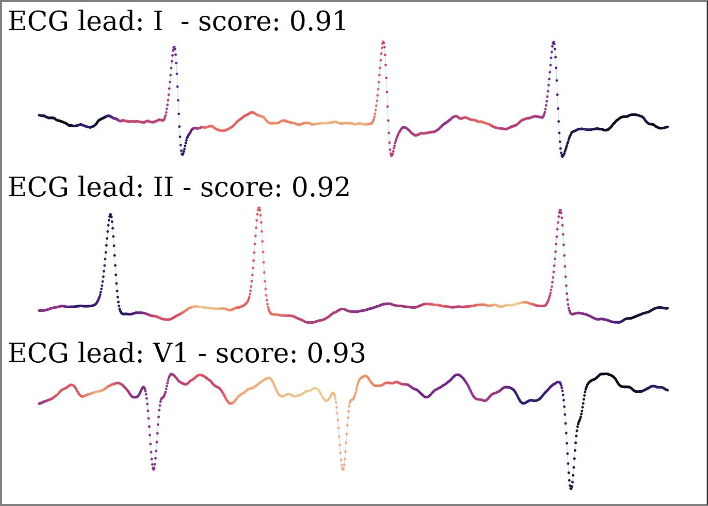}
    \caption{The explanation for another AF ECG recording.}
    \label{fig:AF_2}
\end{figure}
\\
\textit{3) IAVB (first-degree atrioventricular block)}
\begin{figure}[ht]
    \centering
    \includegraphics[keepaspectratio, width=225.53pt]{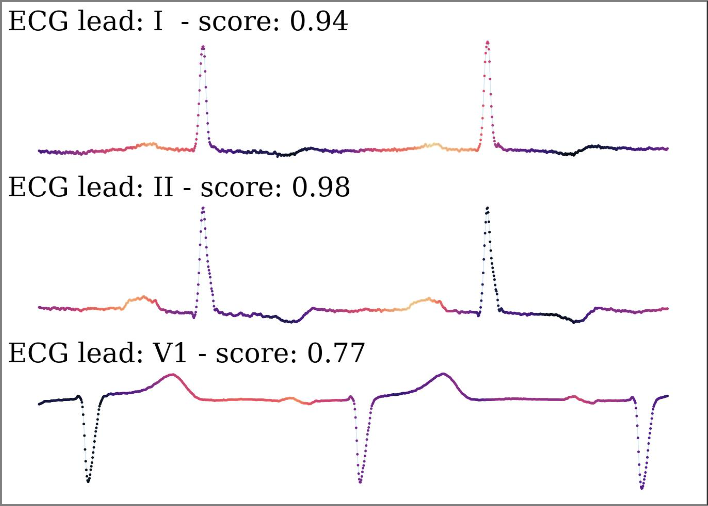}
    \caption{The explanation for another IAVB ECG recording.}
    \label{fig:IAVB_2}
\end{figure}
\clearpage
\textit{4) LBBB (left bundle branch block)}
\begin{figure}[ht]
    \centering
    \includegraphics[keepaspectratio, width=225.53pt]{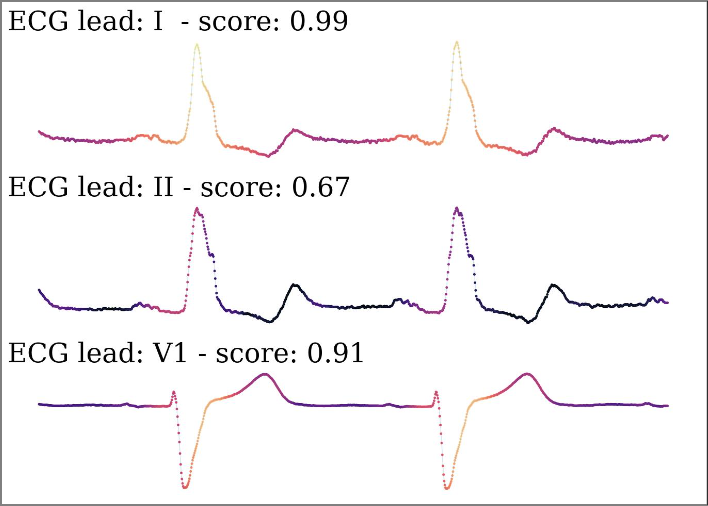}
    \caption{The explanation for another LBBB ECG recording.}
    \label{fig:LBBB_2}
\end{figure}
\\
\textit{5) RBBB (right bundle branch block)}
\begin{figure}[ht]
    \centering
    \includegraphics[keepaspectratio, width=225.53pt]{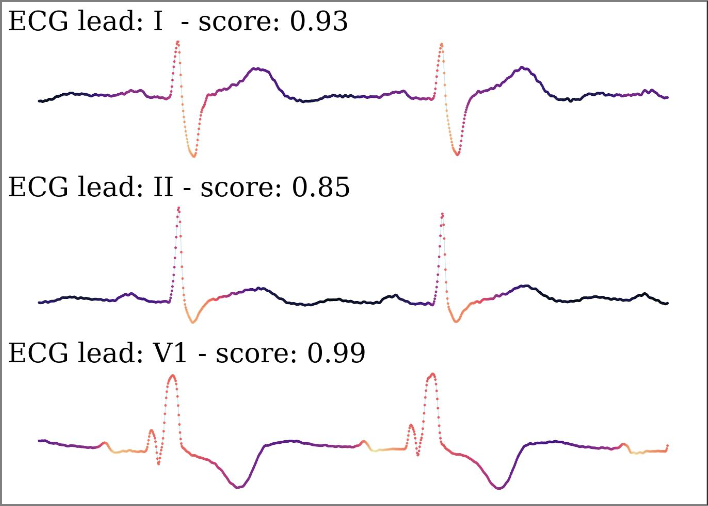}
    \caption{The explanation for another RBBB ECG recording.}
    \label{fig:RBBB_2}
\end{figure}
\\
\textit{6) PAC (premature atrial contraction)}
\begin{figure}[ht]
    \centering
    \includegraphics[keepaspectratio, width=225.53pt]{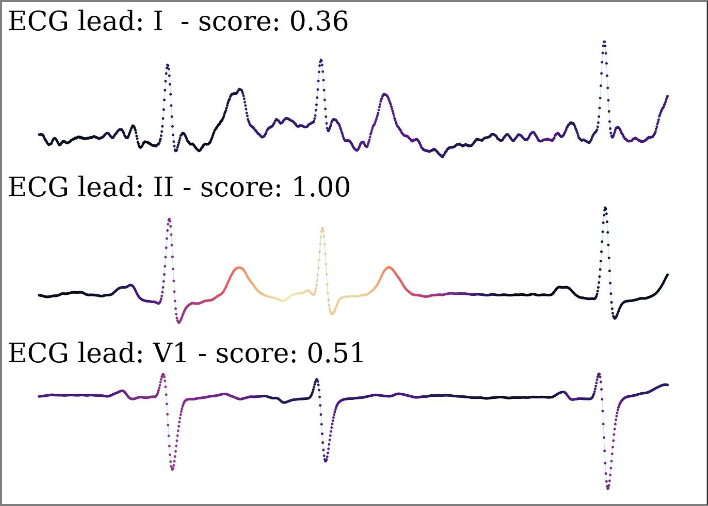}
    \caption{The explanation for another PAC ECG recording.}
    \label{fig:PAC_2}
\end{figure}
\newpage
\textit{7) PVC (premature ventricular contraction)}
\begin{figure}[ht]
    \centering
    \includegraphics[keepaspectratio, width=225.53pt]{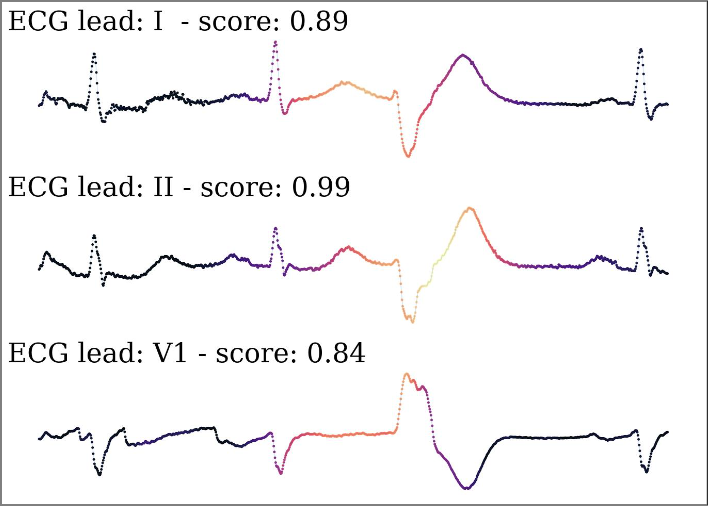}
    \caption{The explanation for another PVC ECG recording.}
    \label{fig:PVC_2}
\end{figure}
\\
\textit{8) STD (ST-segment depression)}
\begin{figure}[ht]
    \centering
    \includegraphics[keepaspectratio, width=225.53pt]{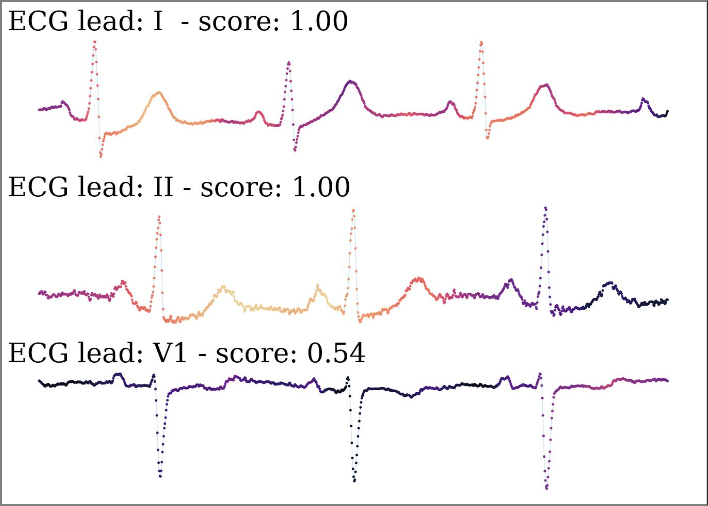}
    \caption{The explanation for another STD ECG recording.}
    \label{fig:STD_2}
\end{figure}
\\
\textit{9) STE (ST-segment elevation)}
\begin{figure}[ht]
    \centering
    \includegraphics[keepaspectratio, width=225.53pt]{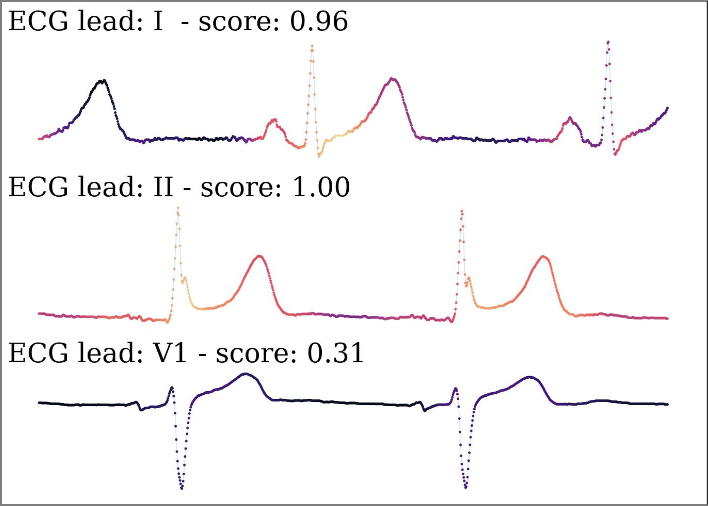}
    \caption{The explanation for another STE ECG recording.}
    \label{fig:STE_2}
\end{figure}

\clearpage
\onecolumn
\subsection{Sanity check}
\begin{figure}[ht]
    \centering
    \includegraphics[keepaspectratio, width=490pt]{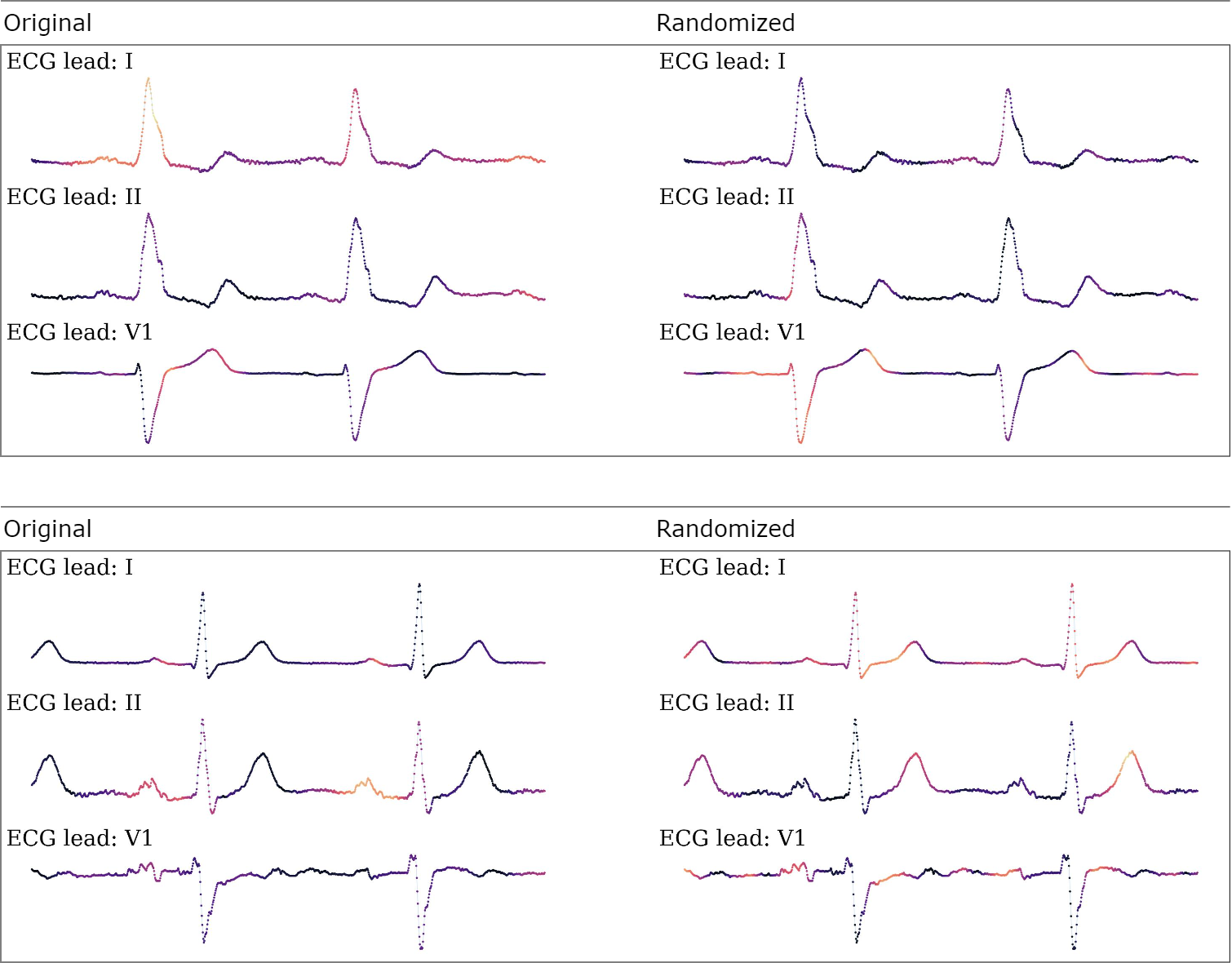}
    \caption{Some examples of comparison between explanations from the original system and the randomized system.}
    \label{fig:Sanity_2}
\end{figure}

\end{document}